**Title**
GAME: Generalized deep learning model towards multimodal data integration for early screening of adolescent mental disorders

**Authors**
Zhicheng Du[1,2], Chenyao Jiang[1,2], Xi Yuan[1,2], Shiyao Zhai[1,2], Zhengyang Lei[1,2], Shuyue Ma[1,2], Yang Liu[2], Qihui Ye[2], Chufan Xiao[1,2], Qiming Huang[3], Ming Xu[3,4], Dongmei Yu[5], Peiwu Qin[1,2,*]

**Affiliations**
[1]Tsinghua-Berkeley Shenzhen Institute, Tsinghua Shenzhen International Graduate School, Tsinghua University, Shenzhen 518055, China
[2]Institute of Biopharmaceutics and Health Engineering, Tsinghua Shenzhen International Graduate School, Shenzhen, Guangdong Province, 518055, China
[3]Shenzhen ZNV Technology Co., Ltd., Shenzhen, 518000, China
[4]Department of Automation, Tsinghua University, 100084, Beijing, China
[5]School of Mechanical, Electrical & Information Engineering, Shandong University, Weihai, Shandong, 264209, China.

**Abstract**
The timely identification of mental disorders in adolescents is a global public health challenge. Single factor is difficult to detect the abnormality due to its complex and subtle nature. Additionally, the generalized multimodal **C**omputer-**A**ided **S**creening (**CAS**) systems with interactive robots for adolescent mental disorders are not available. Here, we design an android application with mini-games and chat recording deployed in a portable robot to screen 3,783 middle school students and construct the multimodal screening dataset, including facial images, physiological signs, voice recordings, and textual transcripts. We develop a model called **GAME** (**G**eneralized Model with **A**ttention and **M**ultimodal **E**mbraceNet) with novel attention mechanism that integrates cross-modal features into the model. GAME evaluates adolescent mental conditions with high accuracy (73.34% – 92.77%) and F1-Score (71.32% – 91.06%). We find each modality contributes dynamically to the mental disorders screening and comorbidities among various mental disorders, indicating the feasibility of explainable model. This study provides a system capable of acquiring multimodal information and constructs a generalized multimodal integration algorithm with novel attention mechanisms for the early screening of adolescent mental disorders.

Keywords: Adolescent mental disorder, Mental health screening, Multimodal learning, Human-Computer interaction, Computer-aided screening.

**Main**
Adolescence is a crucial period of life development during which major psychosocial adjustment takes place[1,2]. A large percentage of mental health disorders that progress into adulthood exhibit symptoms at a young age[3,4], indicating that adolescent mental health issues could degenerate into worse later-life illnesses. Around 13% of adolescents aged 10–19 in the world are diagnosed with different types of mental illness, of which 80 million adolescents aged 10–14 and 86 million adolescents aged 15–19 are deeply affected by mental disorders[5,6]. Especially, ~80% adolescents are unable to receive precise and professional psychological counseling when they demand mental health services[7] and ~50% adolescents with mental disorders have access to psychotherapy[8]. Traditional screening methods for mental disorders include questionnaires and interviews[9], where the results rely on patients' self-reports and psychiatrists' observations[10,11]. However, these methods are inherently susceptible to subjective bias. Furthermore, barriers like stigma in disclosing mental illness or negative attitudes towards professionals[12] lead to inaccurate psychological assessments and a vicious cycle of disease deterioration. To address these limitations, interactive robots providing an enjoyable and acceptable interface with less defensive altitude and hostility offer a promising avenue for unconscious screening[13]. The humanoid robot is more accurate at detecting pediatric mental health problems than parental or child self-reporting[14]. Therefore, imperceptible and interactive screening robot with corresponding algorithm for accurate and opportune screening to adolescent mental disorders can support healthcare agencies and ameliorate the social burden[15,16].

Here, we develop a humanoid robot equipped with well-designed emotional stimuli that facilitates the acquisition of the Multimodal Adolescent Psychological Screening (MAPS) dataset (age 12–15), including facial images, physiological indicators, audio recordings, and textual transcripts (**Fig. 1**). Acquired multimodal dataset are analyzed with statistical model to minimize the distance between prediction and ground-truth provided by screening questionnaires. The Mental Health Inventory of Middle School Students (MMHI-60)[17,18] is a screening questionnaire specially designed to assess Chinese adolescents' mental health and has exhibited high specificity and sensitivity in screening 10 different types of mental disorders (Supplementary **Methods**). We maintain MMHI-60 questionnaire to screen 10 types of mental disorders with additional screening results suggested by experienced psychologists for suicidal tendency. Thus, a total of 12 psychological conditions are labeled as ground truth for individual subject in the dataset, including: (1) depression, (2) interpersonal sensitivity, (3) anxiety, (4) obsessive-compulsive tendencies, (5) paranoid ideation, (6) hostility, (7) academic stress, (8) maladaptation, (9) emotional disturbance, (10) psychological imbalance, (11) suicidal tendency, and (12) overall mental health status[19].

Robotic platforms with human-computer interaction have been utilized for intervention in adolescent mental health[20-22]. However, existing systems lack a computer-aided screening (CAS) algorithm for psychometrics, and the CAS approach has shown promise in the diagnosis of mental disorders in adolescents[23,24]. Models that can process different types of input data (i.e., physical activity, sociability, device usage patterns, etc.) collected from various sensors[25,26] are utilized to recognize specific mental disorders including depression, anxiety, and stress[27-32]. However, the current CAS models employing single-modal feature encounter limitations in constructing a comprehensive representation of the latent multimodal feature space[33], which weakens their performance. Multimodal CAS models have been used to predict psychological disorders and mental states by feature importance ranking, feature selection, and feature concatenation strategies[34-37]. Nevertheless, the screening of specific psychiatric disorders and the lack of interpretability of these models have hindered the adoption of CAS models in clinical applications. Limited exploration exists on whether a generalized model with interpretability could accurately screen adolescents' mental disorders. Therefore, achieving both generalization and interpretability in the CAS system remains a challenge for clinical utility.

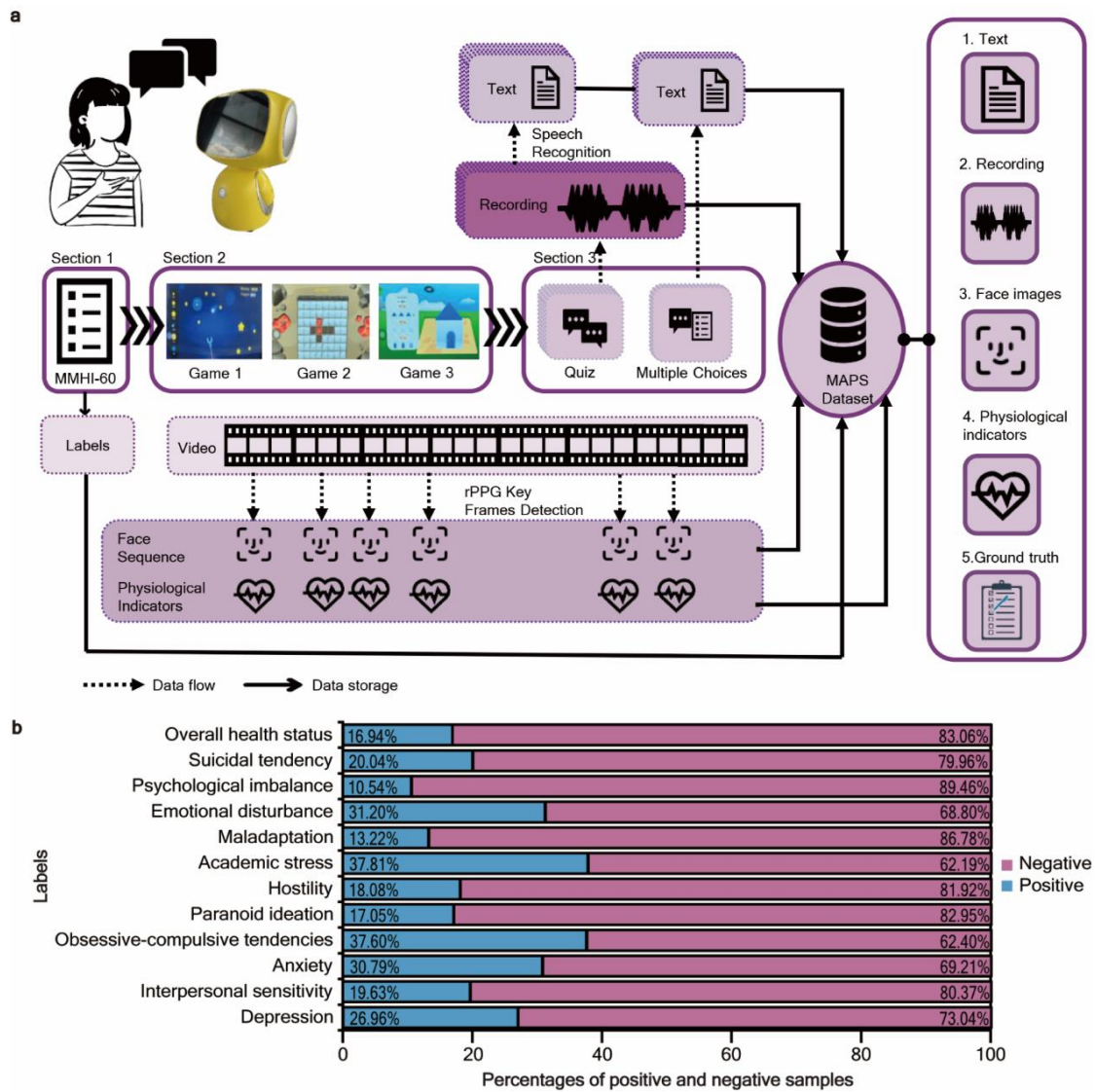

**Figure 1. MAPS data acquisition and database construction. a**, The flowchart of data acquisition. A humanoid robot with a customer-designed Android application that can interact with participants is used for data collection. The data collection procedure has three consecutive sections: psychological screening, emotional stimuli games, and questions-and-answers. The mental health inventory is designed by psychological expert as labels of ground truth. Using remote photo-plethysmography (rPPG) and available processing algorithm, key frames (i.e., images with clear and unmasked face) and physiological indicators are extracted from videos captured during the games and questions-and-answers sections. The responses in the questions-and-answers sections are recorded and converted into text using the speech recognition technique of Iflyrec (https://www.iflyrec.com), supplied by iFlytek. **b**, the sample distributions for 11 types of screened mental disorders and overall mental health status. The ratios depict the imbalance of the MAPS dataset, and the positive samples labeled as 'Overall mental health status' represent the abnormal adolescents.

Hence, we propose GAME (Generalized model with Attention and Multimodal EmbraceNet), a generalized model based on distance-weighted attention mechanisms and multimodal feature fusion in the EmbraceNet backbone network[38] (**Fig. 2**) for adolescent mental disorders screening. GAME extracts eight single-modal features named Expression, Expression nuance, and Eye movement from face images; Physiological signs; MFCC and Wav2vec from audio recordings; PERT and RoBERTa from textual transcripts, respectively. Inspired by psychologists' diagnosing strategy through multifaceted response from adolescents in structured diagnostic and screening interview[39], we propose a novel attention mechanism for multi-scale feature to integrate inter-model correlation weights and eight single-modal features. Cross-modal features named Relation graph and Attention extract deeper information and alleviate the interference of noisy features. Hyper-emotion theory[40,41] indicates that adolescents suffered from mental disorders have abnormal multimodal emotional and behavioral responses to the same interactive stimuli in contrast to healthy subjects. GAME, guided by the hyper-emotion theory, accurately predicts overall mental health status and identifies 11 types of adolescent mental disorders based on multimodal responses. We use GAME to predict comorbidities among adolescents with multiple mental disorders and compare the findings with relevant studies. The ablation experiment that removes one modal input each step and the fusion analysis that evaluates contribution ratio of each model from trained GAME parameters confirm the validity of the modal features and the interpretability of the multimodal fusion.

In summary, this study develops a cost-effective and precise screening robot platform along with GAME to screen early mental illness among adolescents. A practical and adolescent-friendly mental health screening system with accurate and interpretable results will make it possible for CAS systems to be used in clinical settings. The theory-consistent comorbidity prediction demonstrate the GAME's reliability for predicting comorbidity from data-driven perspective. GAME identifies the dominated feature for certain mental disorder and guides the screening design when single-modal data is available, which recommends the clinician pays attention to critical features and directs researchers to implicit patterns or theories found through a data perspective.

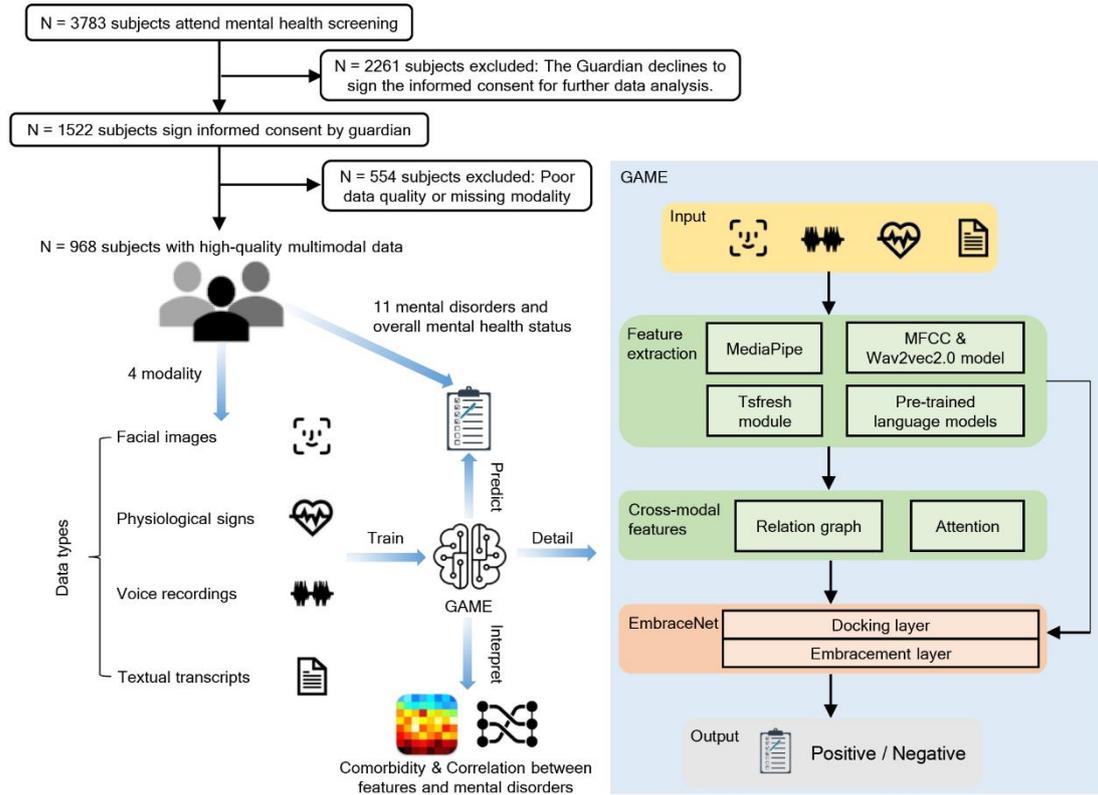

**Figure 2. Pipeline of data processing and GAME's structure.** A total of 3,787 people participated in the mental health screening, retaining 968 samples after exclusion. Based on four types of input, GAME has been trained to predict mental disorders, mining comorbidity and correlation between multimodal features and mental disorders in adolescent. MediaPipe, Mel-Frequency Cepstrum Coefficients (MFCC), Wav2vec2.0, Tsfresh module, pre-trained language models including Robustly Optimized BERT approach (RoBERTa), Pre-training BERT with Permuted Language Model (PERT) are used to extract single-modal features from facial images, voice recording, physiological indicators, and textual transcripts respectively. The extracted features undergo task-level fusion, and then two cross-modal features are generated through unimodal features. Eight single-modal and two cross-modal features are fused by EmbraceNet. BERT means Bidirectional Encoder Representations from Transformers.

## Results
### Multimodal database construction
We construct MAPS dataset with 3,787 Chinese middle school students aged 12 to 15 and filter to 968 (**Fig. 2** and Supplementary **Method**), encompassing four data modalities and 11 adolescent psychological disorders and overall mental health status. The 12 mental health conditions in the dataset have different distribution and the imbalanced positive-to-negative ratios (**Fig. 1b**), which are ranked from high to low as follows: obsessive-compulsive tendencies (6.56), interpersonal sensitivity (5.31), overall mental health status (4.90), academic stress (4.87), hostility (4.53), psychological imbalance (4.09), suicidal tendency (2.71), depression (2.44), emotional disturbance (2.25), anxiety (2.21), maladaptation (1.66), paranoid ideation (1.64). The subjects distribute across multi-centers and cities in Guangdong Province, China.

MAPS collects comprehensive features via portable screening platform compared to the public mental disorder dataset. The IMAGEN study[42] and the Adolescent Brain Cognitive Development Study (ABCD)[43] are large multimodal adolescent mental health datasets, including MRI neuroimaging, behavior, cognition, etc. There are also private clinical datasets that have been used to train AI models for the diagnosis of specific adolescent psychiatric disorders. However, the current datasets are not compatible with portable screening for mental disorders due to high cost and complex data acquisition. MAPS uses an easily accessible and inexpensive data collection platform which can be scaled up for large population screening (Supplementary **Table 2**).

**Attention mechanism and multimodal integration**

With extracted single-modal and cross-modal features, we compare reported machine learning (ML)[44,45] models used for mental disorders diagnosis[46-48], including Support Vector Machine with Polynomial Kernel (SVM-Poly) and Radial Basis Function (SVM-RBF) Kernel, Random Forest (RF), and Gradient-Boosting Decision Tree (GBDT) with GAME, to evaluate the prediction accuracy for 12 mental conditions and robustness of GAME. The performance and stability of the models are assessed using the evaluation metrics of accuracy and weighted F1-Score. Meanwhile, we implement the 10-fold stratified cross-validation (**Fig. 1b**) instead of the random split to evaluate the model's performance. GAME averagely enhances the accuracy of 3.31% - 76.24% (SVM-RBF), 3.31% - 76.55% (SVM-Poly), 3.31% - 15.49% (RF), and 3.93% - 17.98% (GBDT) in comparison to the bracket's baseline models (**Table 2**). In terms of model robustness, GAME enhances the weighted F1-score of the SVM-RBF, SVM-Poly, RF, and GBDT models by 5.07% - 83.31%, 6.57% - 83.94%, 6.34% - 23.78%, and 6.08% - 22.87%, respectively. The wide range of improvements in accuracy and F1-Score indicates the capability of GAME in mental disorders screening.

Table 2 | Models evaluation and comparison for 12 different prediction tasks.

| Ground truth | Evaluation metric | SVM-RBF | SVM-Poly | RF | GBDT | GAME |
|---|---|---|---|---|---|---|
| Overall mental health status | Accuracy | 70.15% (84.30%, 49.17%) | 64.99% (83.06%, 30.68%) | 83.08% (84.40%, 81.61%) | 82.23% (83.68%, 80.99%) | 89.26% (92.78%, 87.63%) |
| | F1-Score | 69.78% (79.47%, 52.87%) | 64.70% (79.49%, 31.32%) | 76.82% (79.21%, 75.37%) | 76.86% (79.09%, 74.84%) | 87.49% (91.92%, 85.42%) |
| Depression | Accuracy | 60.98% (74.38%, 27.38%) | 59.38% (74.38%, 31.50%) | 72.86% (73.45%, 71.59%) | 71.30% (73.14%, 69.94%) | 80.16% (82.47%, 78.13%) |
| | F1-Score | 56.87% (66.04%, 12.49%) | 55.29% (66.97%, 16.68%) | 63.35% (65.12%, 61.89%) | 64.19% (66.32%, 61.61%) | 76.80% (79.15%, 74.00%) |
| Interpersonal | Accuracy | 70.29% | 66.16% | 80.15% | 79.07% | 85.85% |
| sensitivity | | (80.99%, 56.42%) | (80.37%, 41.31%) | (80.58%, 79.03%) | (80.27%, 78.41%) | (88.66%, 83.33%) |
| | F1-Score | 68.56% (75.28%, 60.59%) | 64.68% (74.86%, 42.94%) | 72.59% (74.26%, 71.63%) | 72.88% (74.43%, 71.12%) | 82.76% (86.72%, 79.37%) |
| Anxiety | Accuracy | 57.04% (70.46%, 31.20%) | 54.08% (70.56%, 30.89%) | 68.38% (68.91%, 66.84%) | 66.44% (67.98%, 64.78%) | 77.58% (80.21%, 75.26%) |
| | F1-Score | 52.01% (61.63%, 15.53%) | 49.38% (62.42%, 14.89%) | 58.21% (61.40%, 56.47%) | 59.49% (61.81%, 56.71%) | 74.83% (79.18%, 72.08%) |
| Obsessive-compulsive tendencies | Accuracy | 53.67% (63.95%, 38.22%) | 51.68% (62.60%, 37.91%) | 60.87% (62.71%, 57.65%) | 59.25% (61.68%, 55.90%) | 73.04% (76.04%, 70.10%) |
| | F1-Score | 49.44% (58.00%, 21.87%) | 46.41% (55.40%, 21.21%) | 52.50% (56.97%, 48.60%) | 54.89% (57.14%, 51.63%) | 71.32% (75.17%, 67.85%) |
| Paranoid ideation | Accuracy | 72.17% (82.95%, 46.48%) | 64.69% (82.96%, 31.20%) | 82.58% (83.06%, 81.91%) | 81.38% (82.33%, 80.57%) | 87.08% (88.66%, 85.42%) |
| | F1-Score | 70.13% (78.28%, 49.79%) | 63.74% (75.82%, 32.21%) | 75.71% (76.59%, 75.18%) | 75.74% (76.85%, 74.48) | 83.59% (86.92%, 80.33%) |
| Hostility | Accuracy | 70.95% (82.74%, 51.13%) | 64.81% (82.13%, 28.00%) | 81.47% (81.92%, 80.37%) | 80.41% (81.20%, 79.34%) | 86.78% (88.54%, 84.38%) |
| | F1-Score | 69.21% (77.25%, 55.57%) | 63.66% (76.93%, 26.59%) | 74.24% (75.82%, 73.37%) | 74.50% (75.90%, 72.78%) | 83.54% (86.78%, 78.88%) |
| Academic stress | Accuracy | 57.10% (64.88%, 38.12%) | 54.67% (64.57%, 37.91%) | 61.11% (63.64%, 58.69%) | 59.53% (61.99%, 56.20%) | 74.18% (81.25%, 69.07%) |
| | F1-Score | 49.52% (57.32%, 21.58%) | 47.82% (56.60%, 21.15%) | 53.02% (56.13%, 49.27%) | 54.98% (56.90%, 50.19%) | 73.06% (80.46%, 67.48%) |
| Maladaptation | Accuracy | 68.16% (86.36%, 13.84%) | 62.79% (86.78%, 13.53%) | 86.30% (86.78%, 85.22%) | 84.83% (85.33%, 83.99%) | 90.08% (91.67%, 89.58%) |
| | F1-Score | 67.17% (82.58%, 4.33%) | 62.77% (80.64%, 3.71%) | 80.67% (81.31%, 79.98%) | 80.32% (80.92%, 79.83%) | 87.65% (89.77%, 86.16%) |
| Emotional disturbance | Accuracy | 55.67% (70.35%, 31.61%) | 53.06% (68.91%, 31.30%) | 68.74% (70.56%, 67.56%) | 66.61% (69.11%, 63.85%) | 77.17% (80.41%, 72.16%) |
|  | F1-Score | 50.56% (62.67%, 15.88%) | 48.23% (62.51%, 15.23%) | 59.22% (62.43%, 56.09%) | 59.70% (62.36%, 56.34%) | 73.00% (77.92%, 67.01%) |
| Psychological imbalance | Accuracy | 75.15% (89.46%, 51.44%) | 70.49% (89.46%, 26.96%) | 89.25% (89.46%, 88.22%) | 88.25% (88.84%, 86.88%) | 92.77% (94.79%, 91.75%) |
|  | F1-Score | 76.19% (85.99%, 60.09%) | 71.65% (84.49%, 31.66%) | 84.44% (84.57%, 84.13%) | 84.43% (84.98%, 83.30%) | 91.06% (94.38%, 89.18%) |
| Suicidal tendency | Accuracy | 68.45% (80.06%, 46.77%) | 69.46% (79.96%, 50.91%) | 79.53% (79.96%, 78.20%) | 78.20% (79.34%, 76.24%) | 85.43% (88.66%, 83.51%) |
|  | F1-Score | 65.71% (73.90%, 46.34%) | 66.44% (72.58%, 51.30%) | 71.27% (71.81%, 70.85%) | 71.70% (73.48%, 70.20%) | 82.20% (86.72%, 78.27%) |

The outcomes of ML algorithms are the average values of single-modal features and cross-modal features, while the outputs of GAME are the average values assessed by the 10-fold stratified cross-validation method. Data in red denotes the highest value in the row, while data in blue denotes the row's next-highest value. The maximum and minimum values are denoted by the two-tuple results in parentheses.

Specially, we integrate the baseline outcomes of ML algorithms (**Fig. 3**) for contrasting them with GAME in terms of predicting performance for various forms of mental disorders. The results shows that GAME enhances accuracy by 5.8% - 52.78% (Depression), 4.86% - 44.54% (Interpersonal sensitivity), 7.02% - 46.70% (Anxiety), 9.09% - 35.13% (Obsession-compulsive tendencies), 4.03% - 55.89% (Paranoid ideation), 4.03% - 58.78% (Hostility), 9.30% - 36.27% (Academic stress), 3.93% - 76.55% (Maladaptation), 6.61% - 45.87% (Emotional disturbance), 3.31% - 65.81% (Psychological imbalance), 5.37% - 38.66% (Suicidal tendency), and 4.86% - 58.58% (Overall mental health status), while the weighted F1-Score of GAME is boosted by 10.92% - 64.31% (Depression), 7.49% - 39.82% (Interpersonal sen sitivity), 13.68% - 59.94% (Anxiety), 18.53% - 50.11% (Obsessive-compulsive tendencies), 7.95% - 51.39% (Paranoid ideation), 6.29% - 56.95% (Hostility), 19.12% - 51.91% (Academic stress), 5.07% - 83.94% (Maladaptation), 11.75% - 57.77% (Emotional disturbance), 6.57% - 59.40% (Psychological imbalance), 10.54% - 35.86% (Suicidal tendency), and 8.28% - 56.17% (Overall mental health status), respectively. Furthermore, we employ the metrics of weighted precision, weighted recall, and the normalized confusion matrix to evaluate the performance of GAME across several classification tasks (Supplementary **Fig. 10-12**). GAME outperforms ML methods in both binary and multiple classification indicated by various metrics.

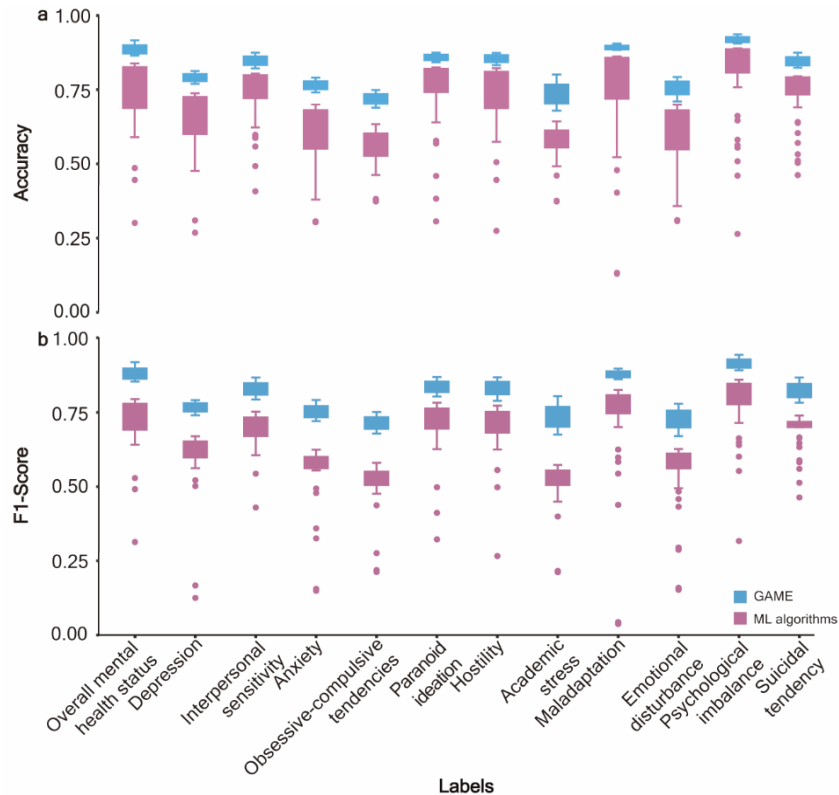

**Figure 3. Evaluation results of comparison between GAME and ML algorithms in various mental disorders. a,** the results assessed by the accuracy in order to evaluate the performance of GAME and ML algorithms work in predicting various types of mental disorders, while the values of ML algorithms are incorporated in accordance with those distinct types of mental disorders. **b,** the results evaluated by weighted F1-score for purpose of measuring the robustness of models.

**Comorbidity among various mental disorders**

We use correlation analysis to evaluate the comorbidities and relevancy levels among different mental disorders in adolescents (**Fig. 4a**). The findings indicate that: (1) There is a comorbidity between depression and anxiety in young individuals; (2) Adolescents with anxiety are at a high risk of experiencing emotional disturbance; (3) Adolescents who suffer from depression and anxiety tend to experience elevated levels of high academic stress; (4) Adolescents with interpersonal sensitivity disorder are more prone to experiencing emotional disturbance, anxiety, depression, and academic stress, where anxiety and depression are more prevalent; (5) Teenagers with paranoid ideation are more susceptible to anxiety, obsessive-compulsive tendencies, and emotional disturbance; (6) Hostility and maladaptation are associated with higher levels of academic stress and psychological imbalance. There is a correlation between hostility and anxiety. (7) Emotional disturbance occurs at a high percentage, followed by academic stress and obsessive-compulsive tendencies. (8) Suicidal tendencies in adolescents may be influenced more easily by depression, anxiety, academic stress, and emotional disturbance. (Detail analysis is shown in Supplementary **Results**). Co-morbidities or correlations among different mental disorders have been shown in published literature and clinical reports, which indicates our data-driven approaches reach similar conclusions as clinical evidence.

In addition, we observe novel comorbidities via the prediction ability of GAME (**Fig. 4b**). The potential comorbidities are inferred from GAME prediction but are not revealed by correlation analysis, for example: (1) maladaptation and paranoid ideation are closely linked to psychological imbalance; (2) there is a comorbidity between paranoid ideation and hostility as well as maladaptation; (3) there is a comorbidities between suicidal tendency with interpersonal sensitivity and paranoid ideation; (4) emotional disturbance has a comorbidity with interpersonal sensitivity. (Further details in the Supplementary **Results)**. A quantitative measure of the comorbidity between different mental disorders or complex interactions can be estimated with our method. The attention mechanism in this study employs the dual relationship in calculating the feature distance, which can be extended to multiple feature similarities when more data points are available later.

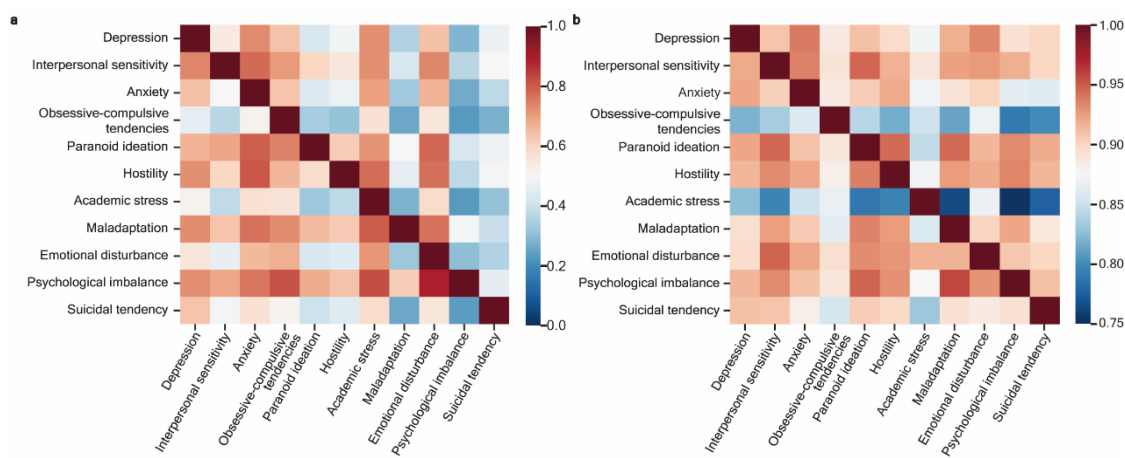

**Figure 4. Comorbidities among 11 different mental disorders in adolescents. a,** the heat map reports the comorbidity association through data statistics. The value of color bar indicates the correlation ratio, which are calculated by the number of samples who are simultaneously suffering from two different mental disorders. **b,** the heat map shows the correlation of GAME predictions. The score of color bars is calculated based on the accuracy obtained from GAME with various model parameters trained by different mental disorders data, with higher accuracy indicating greater resemblance between the two mental disorder. Darker blue indicates poorer correlation while deeper red indicates higher correlation.

**Modality ablation experiments**

Each modal feature can boost the GAME's accuracy in predicting various mental disorder (**Fig. 5a**). The impact of different modal features on the performance of GAME varies, with some exerting stronger influence than others, which facilitates GAME's ability to explain the individual modal's contribution to certain mental disorders predictions. The modal features, ranked from highest to lowest in terms of their contribution to the model's accuracy, are as follows: Wav2vec, Expression, RoBERTa, Expression nuance, Relation graph, Eye movement, PERT, Attention, Physiological signs, and MFCC. The absence of specific modal features can result in a considerable decline in the prediction accuracy of GAME when predicting specific mental disorders, such as Attention features and obsessive-compulsive tendencies, Wav2vec features and emotional disturbance, expression features and academic stress. In terms of weighted F1-score (**Fig. 5b**), the average contribution of modal features to the robustness and stability of GAME is listed in descending order: Attention, RoBERTa, Expression, PERT, Eye movement, Wav2vec, Expression nuance, Physiological signs, Relation graph, and MFCC. Analogously, the removal of certain modal features can greatly diminish the robustness of GAME; for example, Expressions, Physiological signs, Wav2vec, Roberta, and Attention facilitate GAME's stability in predicting anxiety. In addition, Attention and Wav2vec help GAME improve accuracy and robustness in the tasks of screening obsessive-compulsive tendencies and emotional disturbance. The results explainably demonstrate the ranking importance of various factors in mental disorder prediction.

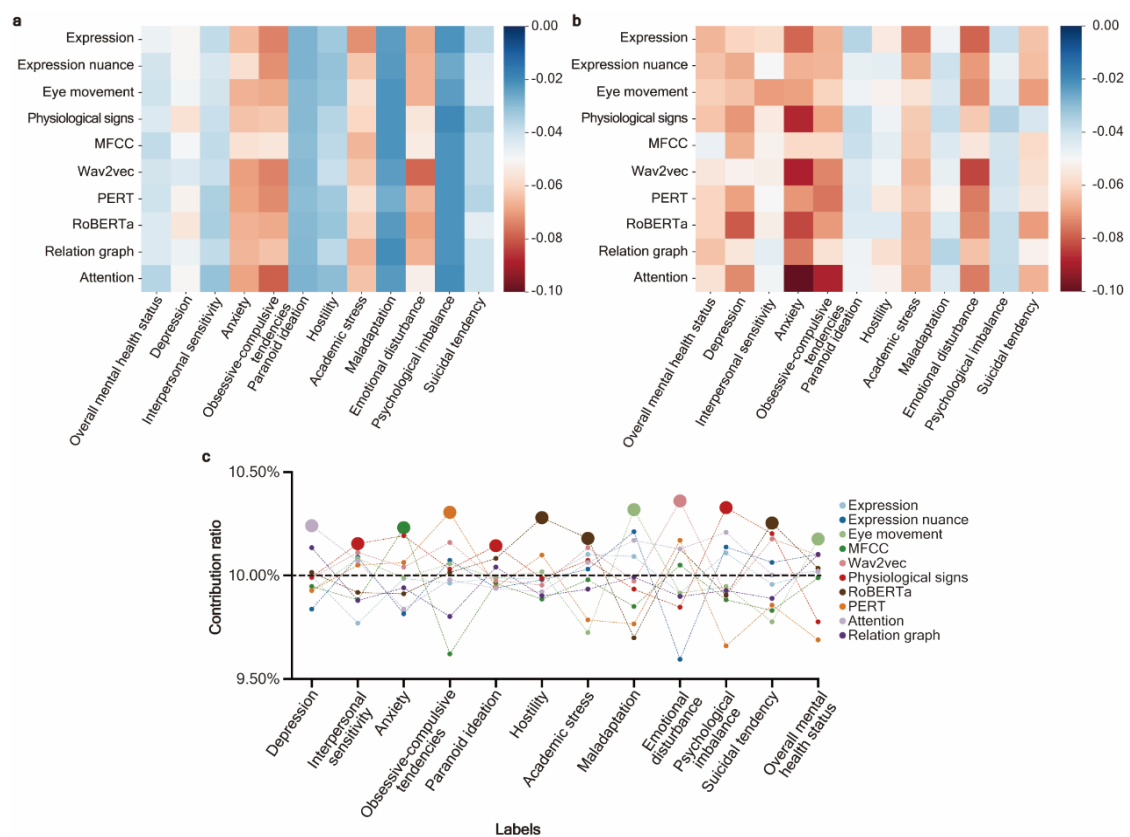

**Figure 5. Ablation and contribution ratio for different modal features. a,** the heat map shows the impact of modal feature elimination on prediction accuracy of GAME. The score of color bar indicates the percentage of accuracy decrease and the symbol '-' represents decline. b, the influence on weighted F1-Score after removing certain modal feature of GAME. Deeper red denotes better correlation, while darker blue suggests lower correlation. These results reflect the relationship between mental disorders in adolescent and corresponding modal features utilized by GAME to predict these disorders. c, the line chart describes the contribution ratio of different features in various GAME prediction tasks, which provides the interpretation of the reasoning why GAME provides this screening decision.

**Modal feature contributions**

GAME indicates the dynamic contribution of each modal feature throughout the multimodal feature fusion to tailor the needs of different scenarios, implying that the same modal feature contribute differently to the prediction of different mental disorders (**Fig. 5c**). We find the following associations between mental disorders and their most important diagnostic features: Attention and Depression; Physiological signs and Interpersonal sensitivity; MFCC (i.e., voice recording) and Anxiety; PERT (i.e., textual transcripts) and Obsession-compulsive tendencies; Physiological signs and Paranoid ideation; RoBERTa (i.e., textual transcripts) and Hostility; RoBERTa and Academic stress; Eye movement and Maladaptation; Wav2vec (i.e., voice recording) and Emotional disturbance; Physiological signs and Psychological imbalance; RoBERTa and Suicidal tendency; as well as Eye movement and Overall mental health status. These findings explain the deterministic features utilized by GAME to make predictions for certain mental disorders, which are consistent with the screening methods used in previous work[49,50] (Detailed analysis in Supplementary **Results)**. Under resource- or time-limiting scenarios, the conclusion about important feature provides guidance for choosing the most valuable modality for certain mental disorder screening.

**Discussion**

CAS models for biomedical applications[51] have experienced rapid development[52-55] and multimodal learning has attracted attention for the screening and diagnosis of multiple diseases[56,57]. Nevertheless, the absence of screening hardware slows down the progress of CAS in psychology and restricts the creation of a generalized and interpretable multimodal CAS for screening adolescent mental disorders. To resolve this problem, we design and create an interactive robot with a well-designed Android APP to screen adolescent disorders unconsciously in a large population. Then we build a MAPS database and develop a generalized multimodal model, named as GAME, to predict adolescent mental disorders with high accuracy and stability. The integration of multiple feedback features is a promising predictor of psychological disorders in adolescents.

The multimodal feature fusion and the attention mechanism boost the universality of GAME in the task of screening diverse mental disorders, where previous deep learning (DL)[58,59] models are developed specifically for certain mental disorders[60,61]. GAME evaluates adolescent's mental health conditions with an accuracy of 73.34% – 92.77%, a F1-Score of 71.32% – 91.06%, a specificity of 73.24% – 93.14% and a sensitivity of 73.04% – 92.77%. Since other psychometric tools were reported to have ~70% specificity [62,63], GAME is a more effective and powerful tool for screening adolescent mental disorders. Modality ablation shows that each modal feature provides a positive contribution in predicting performance. Notably, the absence of Attention leads to a ~10% reduction in model performance when predicting anxiety and obsession-compulsive tendencies. In a nutshell, GAME is superior to conventional ML algorithms and screening tools in prediction performance due to its thorough feature extraction and cross-modal information mining.

Comorbidity is a common phenotype for those who have mental illnesses[64].

Adolescents with mental disorders require comorbidity analysis to create a precise psychological portrait. Comorbidities has clinical implications for the diagnosis of mental disorders, the prescription of appropriate treatments, and the long-term management[65]. However, to the best of our knowledge, few researchers utilize multimodal algorithms to mine comorbidities among adolescent psychological disorders. GAME can quantify the relevancy magnitude between different mental disorders in adolescents, which improves the accuracy of the mental disorders screening and provides insights for development of adolescent psychological theories through data-driven perspective. For example, GAME predicts a comorbidity between emotional disturbance and interpersonal sensitivity, shown in empirical research[66], which indicates that unstable social relationships cause emotional disorders. GAME as a digital assistant to prompt the psychiatrist to give priority to the interpersonal sensitivity rather than emotional disturbance. The GAME can be extended to discover novel comorbidities if more modal features and mental disorder types are provided.

Interpretability is crucial for the development and application of CAS systems in clinical settings. Unexplained or opaque models (known as "black boxes") make it difficult to understand the logic reasoning of clinical decision[67]. By dissecting the trained GAME's parameters, we explain how GAME makes predictions through the contribution ratio for each modal feature during diverse prediction tasks, which uncovers the relationship between mental disorders and modal features through modality ablation. For example, GAME suggests that Physiological signs is more important than other modal features in predicting interpersonal sensitivity, which is consistent with the report that interpersonal sensitivity is associated with higher systolic blood pressure[68]. GAME guides future research directions through comorbidity relationships and correlation between features and mental disorders. For instance, GAME predicts that maladjustment and paranoid ideation are possibly linked to psychological imbalance. However, there is currently no relevant work to show the comorbidity between them, and future work is required to fill this gap.

There are still limitations in this study. First, even that GAME has been validated, the size of the MAPS dataset is modest, which restricts the performance of data-driven models and necessitates the collection of larger samples to enable GAME to learn subtle features about adolescent mental disorders. Adolescents' mental disorders are closely related to their living environment[69]. In the future, we can enlarge the MAPS dataset to include more cities and countries with diverse economical stages, geographical environments, and social culture. Second, the materials of emotional stimuli may not be abundant enough. To improve the reliability of audiovisual stimuli[70,71], emotionally elicited film clips should be included. Third, public multimodal datasets can be used to train GAME for widespread applications. However, multimodal datasets for screening of adolescent mental disorder are not available. Transfer learning with a pre-trained model can be adopted to extra psychometric applications instead of screening. Fourth, GAME can be used to tackle the issue of modalities absence, which has not been addressed in computational psychology. Real-world datasets often contain inadequate modality data for a variety of reasons, like data privacy, failed acquisitions, data corruption, and costly testing[72]. However, the missing modality problem has been studied in other diseases' diagnosis[73].

In summary, this study demonstrates that a cheap (< $400), portable, interactive, expansible robot with vivid emotional stimulation materials can effectively facilitate screening and diagnosis of adolescent mental health disorders. GAME with theoretical support has the advantages of high accuracy, strong stability, and interpretability, which meets the needs of adolescents' mental disorder screening and unveil the relationship among various mental disorders as well as the correlation between mental disorders and modalities from a model-driven perspective.

## Methods
The study was approved by the Office of Research Ethics of the Tsinghua University, Shenzhen International Graduate School protocol No.41 in 2021.

### Design of Android application
The system of data transfer and the database management are developed based on Spring Boost 2.0, Spring Cloud, Mysql, VUE, Docker, Remote Dictionary Server (REDIS), and EQUEUE technologies, etc. There are two phases in the development of an Android application: screening protocol design and code development. Firstly, we collaborate and consult with professional psychologists, psychological counselor from middle school, and representative parents to identify the requirements and appropriate tools for adolescent mental disorders screening. Then, we formulate the interaction scheme and functional architecture of the application. Once we validate the engineering feasibility of the scheme and structure, we proceed with designing the user interface (UI) and user experience (UE). We follow the code development order of application (APP) client, application programming interface (API) server, and background database management system. In detail, we use Java and the front-end framework VUE for development of the application client, employ Restful API and Domain-driven Design (DDD) technologies for application API server development, and utilize REDIS and MySQL for background database management systems. Upon completing the application development, we conduct application program testing, including App content testing, App performance testing, App function testing, App visual testing, debugging, and repairing bugs. Finally, we deploy the application onto the interactive robot for on-site screening (Supplementary **Fig. 1–9**). The screening platform we develop provides objective and involuntary screening appropriate for repetitive screening, whereas questionnaire screening has a noticeable bias for recurrent screening. Also, the APP's content facilitates personalized further development, allowing researchers to tailor different stimulus materials and meet the various demands of psychological screening and diagnosis.

### MAPS Dataset Collection
Our adolescent multimodal mental health screening dataset contains facial, textual, acoustic, and physiological data, four data modals, which are collected from multiple middle schools in Guangdong Province with 3783 volunteers ranging from 12 to 15 years old and filtered to 968 after exclusion (Supplementary **Methods**). Each data is collected by a humanoid robot. The main components of this robot include a touch screen, a camera, a speaker, and a recording device. The touch screen displays the test content and allows interaction with the test taker. The camera records video of the volunteers' faces, and the recording device records the volunteers' voices during the test. The recorded data is transferred to a configured personal computer for storage. An Android app installed in the robot system completes the entire testing and data collection process (Supplementary **Methods**). Personal information, such as gender, age, class number, and student ID, is required prior to data collection. The volunteer will enter all of the above information into the robot via the touch screen. The recorded video of the acquisition process and classroom environment is provided in the Supplementary **Videos** and Supplementary **Fig. 13**.

To minimize the physical and psychological discomfort experienced by adolescent participants during screening caused by a wearable device, we use a high-resolution camera installed in the robot to collect video data and calculate physiological signs by the rPPG algorithm integrated in the back-end server. The rPPG[74] algorithm, coined as non-contact PPG[75,76], is a technique to analyze the face video to extract physiological indicators, including heart rate, heart rate variability, changes in blood pressure, and respiration rate. Stress and relaxation levels can be calculated using a DL algorithm and the arousal-valence emotion model[77,78] based on physiological indicators. Eventually, we obtain six physiological metrics and save them in the database. The volunteer may move significantly during the screening process, potentially causing the rPPG algorithm to fail at deriving certain physiological indicators. Only the key and clear frames in the videos identified by the rPPG algorithm can be used to acquire the physiological indicators, and we save the pairs of face images and physiological signs to maintain a consistent correspondence between them.

**MMHI-60**

The MMHI-60 is adapted from the Symptom Checklist-90 (SCL-90)[79], which was designed through a two-year follow-up survey on the mental problems of middle school students in more than 100 schools across China and has been successfully applied to the mental disorders screening for Chinese middle school students[80]. The MMHI-60 consists of 60 questions to measure relevant symptoms of 10 distinct mental problems (including depression, interpersonal sensitivity, anxiety, obsessive-compulsive tendencies, paranoid ideation, hostility, academic stress, maladaptation, emotional disturbance, and psychological imbalance). For each question, the respondent assigns a score ranging from 1 to 5, depending on whether they have recently undergone a specific type of symptom or behavior, which represents none, mild, moderate, heavy, and serious, respectively.[81]. The MMHI-60 uses a 5-point Likert scale, where a score of 2-2.99 indicates the presence of mild problematic symptoms; 3-3.99 suggests moderate symptoms; 4-4.99 indicates the presence of severe symptoms; and a rating of 5 denotes severe psychological symptoms. Final score is the average score of its corresponding questions, allowing the participants to be identified as having the potential for symptoms of a relative mental disorder. The mental health issue is recognized when the average score of the subscale is equal to or higher than 2, which will be regarded as positive. The ground truth of overall mental health status is obtained by combining all the scores from subscales (i.e., the higher the score, the worse the overall mental health status), and the ground truth of suicidal tendency is obtained by both the MMHI-60 and diagnostic advice from the psychiatrist. The question list of the MMHI-60 is presented in the Supplementary **Methods**.

**Theoretical Support**
This work relies on hyper-emotion theory, which supports GAME for the plausibility of predicting psychological conditions based on the magnitude of emotional responses to external stimuli within adolescents. Mental diseases originate with a cognitive appraisal that undergoes a chain of unconsciously transitions leading to a fundamental emotion, such as happy or angry. The hyper-emotion theory contains five principles: (1) The principle of unconscious transitions to fundamental emotions. People develop a series of unconsciously shifts from a physiological sensation or cognitive assessment to a fundamental emotion that is appropriate to the circumstance but aberrant in its response intensity. Such transitions lead to the start of a psychological illness, but they persist during the illness[41]. (2) The principle of no voluntary control. People are unable to control their basic emotions during straightforward cognitive assessments. (3) The ontological principle. The ontogeny of social mammals serves as the foundation for the development of basic emotions, as the source of psychological diseases. (4) The principle of vulnerability. The susceptibility of individuals to psychiatric diseases varies according to intrinsically established conditions and adverse circumstances. (5) The principle of inferential consequences. People pay more attention to an abnormal basic emotion, think about it, and try to identify its causes. They become skilled at making inferences about the topic they are pondering, and their inferences can perpetuate and worsen the mental illness.

In brief, hyper-emotion theory endorses the notion that individuals occasionally perform cognitive assessments, which they may consciously recognize, resulting in an unconscious transition towards a fundamental emotion of heightened intensity. The episode may be brief or it may intensify into a full-fledged psychological disease, depending on individual constitutional and environmental factors. The theoretical foundation of this study is to allow teenagers to show their unconscious emotion perturbation to emotional stimuli from the interactive robot.

**Data Preprocessing**
To ensure that the feature vector dimensions entered into GAME are consistent, we preprocess the recording data as follows to ensure that the length of the recordings is the same for all subjects. We set the valid recording duration to 10 seconds as the average length. If the recording length is longer than the average length, the excess frames will be truncated. If the recording length is less than the average length, it will be padded with zeros. Because the feature extractor can automatically solve the problem of length inconsistency (i.e., inconsistent length of text, face video, and physiological index), we do not need to perform a preprocessing step for the other data modalities.

**Single-modal Feature Extraction**
The purpose of feature extraction is to retain decent separability (e.g., help GAME classify data accurately) and reduce computing costs while mapping the sample from a high-dimensional feature space to a low-dimensional feature space. The followings are the algorithms used to extract single-modal features or cross-modal features.
(1) Feature extraction for audio recordings

Mel-scale Frequency Cepstral Coefficients (MFCC)[82] is used as the feature of acoustic recordings that is commonly used in audio-related tasks like speech recognition and speaker recognition. An audio is subjected to a rapid Fourier transform, Mel filter bank, logarithmic operation, discrete offline transform, and dynamic feature extraction in order to acquire the MFCC feature. We obtain the MFCC feature extracted by speech-features-module (https://github.com/jameslyons/python_speech_features), which is a python package for audio signal processing and audio feature extraction.

The calculation of MFCC can be divided into the following steps: first, frame the signal into brief frames. Under the premise that the audio signal doesn't vary substantially across small time scales, we confine the signal length into 25 ms, which is consistent with the acquisition frequency of 16 Khz, corresponding to $0.025 * 16000 = 400$ frames. We set frame step as 10 ms (160 samples), which allows some overlap between steps. The first 400 sample frame starts at sample 0, the next 400 sample frame starts at sample 160 etc. until the end of the speech file is reached. The second step is to calculate the power spectrum of each frame. One set of 12 MFCC coefficients is retrieved for each frame. Then, the Discrete Fourier Transform (DST) for each frame will be determined using the following formula:
$$S_i(k) = \sum_{n=1}^{N} s_i(n)h(n)e^{-j2\pi kn/N} \quad 1 \leq k \leq K,$$
where $h(n)$ means the analysis window with $N$ samples (i.e., hamming window) and $K$ is the length of the DFT. Additionally, $s(n)$ means time domain signal, whose $i$ ranges over the number of frames. The $S_i(k)$ and $P_i(k)$ implies the time-domain frame and the power spectrum of frame i, respectively. Then, the periodogram-based power spectral estimate for the speech frame $s_i(n)$ is given below:
$$P_i(k) = \frac{1}{N}|S_i(k)|^2.$$
We square the output after taking the complex Fourier transform's absolute value. The next step is to calculate the Mel-spaced filter bank, take the log for each of the 26 output from previous step, and finally take DCT of the 26 log filter bank items to obtain 26 cepstral coefficients. Consistent with traditional automatic speech recognition task settings, we keep the lower 13 of the 26 coefficients as the resulting features.

In addition to the conventional speech recognition algorithm for feature extraction, we also employ the self-supervised pre-training DL model wav2vec 2.0[83] to embed the audio. In contrast to other models, wav2vec 2.0 performs the best in many standard voice tasks[84]. Thus, we employ wav2vec to extract features from audio recordings of adolescents. Wav2vec2.0 encodes speech audio using a multi-layer convolution neural network and subsequently masks portions of the latent speech representations. The model is trained using a contrastive manner in which the real latent is differentiated from fake latent. The latent representations are supplied to a Transformer[85] network to produce contextualized representations.

(2) Feature extraction for textual transcripts

For text data, we use **Ro**bustly optimized **BERT a**pproach (RoBERTa)[86] and PERT[87] to extract the textual feature. The features produced by these two models varied because of the different architectures and the Chinese corpus used for training, so we employ the two models' output as the inputs to improve the robustness and reliability of GAME in predicting adolescents' mental disorders. RoBERTa and PERT are enhanced versions of BERT[88], exhibiting capability in numerous tasks including text classification, machine reading comprehension, and text prediction. Based on pre-trained models, we extract features directly without fine-tuning. RoBERTa is an improved BERT model that can match or exceed the performance of all post-BERT methods and provides a detailed evaluation of the impact of hyper-parameter tuning and change of training set size[86]. PERT is a permuted language model to recover the word orders from a disordered sentence, and the objective of PERT is to predict the position of the original word, which outperforms other BERT variants on a few tasks[87]. The use of PERT and RoBERTa can extract the features of text data from different perspectives.

(3) Feature extraction for facial images

The features of the face images are extracted using MediaPipe FaceMesh[89]. From a single image without depth information, MediaPipe FaceMesh can provide the 3D shape of a human face represented by 468 points with 3D coordinates. We use the pre-trained model to generate the features of each image in the sequence, in which the face is resized to 256 × 256. The image will first be processed by a face detector to mark a rectangle area containing the face and landmarks such as eye centers and nose tips. Then the face rectangle is cropped, resized, and fed to a deep neural network to generate a vector of 3D landmark coordinates.

Furthermore, we use MediaPipe Iris[90] to track the eye movements of the volunteer. After MediaPipe FaceMesh detects the face area and eye landmarks, a DL model is trained to mark subtle positions such as iris position, eye contour, and pupil location. The position of each eye is represented by a pair of coordinates. Eye movement can be utilized to infer users' behavior and cognitive status in human-computer interaction[91], since pupil response is closely related to cognitive and emotional processes[92].

(4) Feature extraction for physiological indicators

Tsfresh[93] is a Python package for extracting features from time series data, which employs 63 methods to obtain features, including absolute energy, the highest absolute value, etc. The Tsfresh module processes the time series data in three stages. The first phase is feature extraction, in which the algorithm characterizes the time series and generates aggregated time series features using the module of feature calculators. Each extracted feature vector is weighted according to its p-values to determine significance in achieving the desired outcome during the feature significance testing phase. The final phase is the multiple test procedure, which determines what features need to be retained[94]. The detailed implementation of feature extraction is described in Supplementary **Methods**.

**Z-Score Normalization**

After extracting the modal features from the individual modality data, we transform them using Z-score normalization to convert the feature vectors into a consistent spatial dimension. The following formula is used to determine the Z-score in statistics:

$$Z = (x - \mu)/\sigma$$

where, $Z$ means Z-score, $x$ is the original value being evaluated, $\mu$ denotes the mean value of all data and $\sigma$ implies the standard deviation. Cross-modal feature extraction and multimodal feature fusion are performed after Z-score normalization.

**Cross-modal Feature Extraction**

From eight single-modal features standardized by Z-score, we extract cross-modal features: Relation graph and Attention. Cross-modal features mine the relationship between various modal features, assisting GAME to use the correlation among modal features to predict a variety of mental disorders. The relationship graph is a weighted undirected graph where each node is a single-modal feature and the weight of an edge is determined by the distance between the linked node's features. Since the length of different unimodal features varies, we apply the Dynamic Time Warping (DTW)[95] approach to compare the similarity between two time series of varying lengths or calculate the distance between them. Consequently, the relation graph has eight nodes in the vertex set and 32 weighted edges in the edge set, which will be represented in an $8 \times 8$ adjacency matrix.

For the calculation process of DTW, suppose we need to measure the distance between two example series $X = \{x_1, x_2, ..., x_m\}$ and $Y = \{y_1, y_2, ..., y_n\}$. We set $M(X, Y)$ as the $m \times n$ point-by-point distance matrix between sequences X and Y, where each point (i, j) is distance calculated by $M_{i,j} = (a_i - b_j)^2$ after the alignment between $x_i$ and $y_j$ due to length variation. The elements of X and Y are mapped along a warping path P to minimize the distance between them and P is a group of index pairs that make up a matrix traversal, which is defined as:

$$P = <(e_1, f_1), (e_2, f_2), ..., (e_s, f_s)>$$

In order to avoid the problem of combinatorically explosive (i.e., examining every possible combination), the following prerequisites must be satisfied for a warping path to be valid: (1) Boundary Condition: $(e_1, f_1) = (1,1)$ and $(e_s, f_s) = (m, n)$, which guarantees that the warping path starts at the beginning of both series and terminates at the endpoints of them. (2) Monotonicity condition: $e_i \leq e_{i+1}, 0 < i \leq m$ and $f_i \leq f_{i+1}, 0 < i \leq n$, which preserves the chronological sequence of points. (3) Continuity condition: $e_{i+1} - e_i \leq 1, 0 < i \leq m$ and $f_{i+1} - f_i \leq 1, 0 < i \leq n$, which restricts the forward transitions to nearby points in next time-stage. We define $\text{dist}(X_{x_i}, Y_{y_i})$ be the distance between elements at point $x_i$ of sequence $X$ and $y_i$ of sequence $Y$. As a consequence, the distance for optimal path P is equal to

$$D_P(X_{x_i}, Y_{y_i}) = \text{dist}(X_{x_i}, Y_{y_i}) + \min\{D_P(X_{x_{i-1}}, Y_{y_i}), D_P(X_{x_i}, Y_{y_{i-1}}), D_P(X_{x_{i-1}}, Y_{y_{i-1}})\}.$$

If we use $\Theta$ to represent the realm of all potential paths and $P^*$ is the shortest warping path. Hence, we can calculate the optimal warping path that
$$P^* = \min_{P \in \Theta}(D_P(X, Y)).$$
Let $p_i = M_{X_{e_i}, Y_{f_i}}$ be the distance between elements at position $e_i$ belong to X and $f_i$ of Y. The DTW distance between two series is obtained by the formula:
$$D_{P^*}(X, Y) = \sum_{i=1}^{S} p_i.$$
An exact solution of the best route $P^*$ can be made using a dynamic programming approach.

With attention mechanism, the model can extract crucial feature, assign each input component a different weight, and reach more precise judgments. Similarly, we leverage the DTW method with attention weights, and the detailed process is described as the following. First, we select one of the single-modal features as the benchmark and use the DTW technique to determine the distance with the other remaining features. We use $d_i$ to denote the distance between any two single-modal features, $d_i = DTW(M)$, $0 \leq i \leq 7$, where M is the feature vector set with eight unimodal features. Second, we utilize the softmax function convert the distance set $D = \{d_i\}$, $0 \leq i \leq 7$ produced in the first step into a weight set $W = \{w_i\}$, $0 \leq i \leq 7$ to satisfy the requirements that $\sum_{i=0}^{7} w_i = 1$. Third, the corresponding feature vector is weighted based on the weight set obtained in the second stage, and the outcome is then added in bitwise to the benchmark feature vector. The addition operation is based on the sequence correspondence in the DTW algorithm, and the dimensionality of the resulting feature vector is the same as the benchmark. Forth, repeat the same procedures using each of the eight single-modal features as the reference to generate eight new feature vectors, and then concatenate them as the attention modal feature.

**Multimodal Feature Fusion and Classification**
(1) Task-level feature fusion
Here we use a simple strategy of averaging all feature vectors including text, audio, and the face landmarks. The average of eight sentence features is used to describe the overall features of the text modality, the average of five audio features is used to describe the features of the audio modality, and the average of multiple face landmarks is used to represent the face's 3D shape feature. For the iris location in the face image, we use it directly without any preprocessing before multimodal fusion.
(2) GAME

GAME extracts eight unimodal features from four individual modality data and creates two novel cross-modal features based on the single-modal features. We then employ EmbraceNet[38] as the backbone network of the multimodal feature fusion method, and the network structure of GAME is shown in **Figure 2**. EmbraceNet is a robust multimodal fusion model allowing for excellent compatibility with any network structure, which considers correlations between various modalities. Additionally, GAME can handle missing data. There are two main parts in EmbraceNet: the docking layers and the embracement layer. Docking layers convert the feature vector of a modality into a format suitable for integration, where the original feature vector is multiplied with parameter matrix and added by bias matrix. For example, suppose that there are m modal features extracted by corresponding network models, the output vector from the $k^{th}$ network model will be called $x^{(k)}$, where $1 \leq k \leq m$. The $i^{th}$ component of the input vector for the $k^{th}$ docking layer is written as

$$z_i^{(k)} = w_i^{(k)} \cdot x^{(k)} + b_i^{(k)},$$

where $w_i^{(k)}$ and $b_i^{(k)}$ are weight and bias vector that correspond to the $k^{th}$ docking layer, respectively. Finally, the output $d^{(k)}$ of the $k^{th}$ docking layer is obtained by applying an activation function $f_a$ to $z_i^{(k)}$, i.e.,

$$d_i^{(k)} = f_a(z_i^{(k)}).$$

All the outputs of the docking layers are vectors with c dimensions, where the hyper-parameter c (embracement size) can be configured if necessary (32 in GAME).

In the embracement layer, the outputs of the docking layers are fused into a vector representing all modal information using a probability-based approach as follows. Consider $r_i = [r_i^{(1)}, r_i^{(2)}, \ldots, r_i^{(m)}]^T, 1 \leq i \leq c$ is a vector obtained from a multinomial distribution, $r_i \sim$ multinomial$(1, p)$, where $p = [p_1, p_2, \ldots p_m]$ and $\sum_{k=1}^{m} p_k = 1$. Only one $r_i$ equals to 1 in accordance with the definition of the multinomial distribution, and all other values are equal to 0. The vector $r^{(k)} = [r_1^{(k)}, r_2^{(k)}, \ldots r_c^{(k)}]^T$ is calculated with the output vector from docking layers $d^{(k)}$ as

$$d'^{(k)} = [d_1'^{(k)}, d_2'^{(k)}, \ldots, d_c'^{(k)}]^T = r^{(k)} \circ d^{(k)},$$

where ∘ means the Hadamard product, which will multiple the elements in bitwise (i.e., $d_i'^{(k)} = r_i^{(k)} \cdot d_i^{(k)}$). Ultimately, the $i^{th}$ element of the output vector belonging to the embracement layer $e = [e_1, e_2, \ldots, e_c]^T$ is determined by the following formula: $e_i = \sum_{k=1}^{m} d_i'^{(k)}$. The terminal network uses it as an input vector and outputs a final category label for the specified classification task.

**Experimental Evaluation Metrics**

In order to comprehensively evaluate the performance of GAME on imbalanced datasets, we implement a stratified k-fold cross-validation approach, where k is set as 10. Accuracy, weighted F1-score, weighted Precision score, weighted Recall score, and normalized confusion matrix are calculated. The accuracy can be computed by the formula:

$$Accuracy = \frac{TP + TN}{TP + TN + FP + FN}$$

The F1-score is calculated by Precision score and Recall score. The definitions of the weighted Precision score and weighted Recall score are listed as the following.

$$Precision_i = \frac{TP_i}{TP_i + FP_i}$$

$$Precision_{\text{weighted}} = \frac{\sum_{i=1}^{L}(Precision_i \times w_i)}{L}$$

$$Recall_i = \frac{TP_i}{TP_i + FN_i}$$

$$Recall_{\text{weighted}} = \frac{\sum_{i=1}^{L}(Recall_i \times w_i)}{L}$$

$$w_i = \frac{Sn_i}{Tn}$$

where i depicts class index, L is the total class number, TP means True positive, TN is True negative, FP represents False negative, FN is False negative, Sn is sample number of specific class, and Tn is the total sample number. The weighted F1-Score can be determined as

$$F1_{\text{weighted}} = 2 \times \frac{Precision_{\text{weighted}} \times Recall_{\text{weighted}}}{Precision_{\text{weighted}} + Recall_{\text{weighted}}}$$

Normalized confusion matrix in cross validation is obtained by averaging each fold of the confusion matrix and then normalizing the output.

**Acknowledgements**

We appreciate the participants in this study for their time and valuable commitment to this study. We thank Dr. Yongjie Zhou from Shenzhen Mental Health Center for her time and comments about the labeling of ground truth for each adolescent mental disorder. This work is supported by funding from the National Natural Science Foundation of China 31970752; Science, Technology, Innovation Commission of Shenzhen Municipality JCYJ20190809180003689, JSGG20200225150707332, JCYJ20220530143014032, ZDSYS20200820165400003, WDZC20200820173710001, WDZC20200821150704001, JSGG20191129110812708; Shenzhen Bay Laboratory Open Funding, SZBL2020090501004; Department of Chemical Engineering-iBHE special cooperation joint fund project, DCE-iBHE-2022-3; Tsinghua Shenzhen International Graduate School Cross-disciplinary Research and Innovation Fund Research Plan, JC2022009; and Bureau of Planning, Land and Resources of Shenzhen Municipality (2022) 207.

**Competing interests**

The authors have declared no competing interests or potential conflicts that could have appeared to influence the work reported in this paper.

**Data availability**

Due to requirements for ethical approval and the possibility of jeopardizing participant privacy, we will publish our dataset after feature extraction instead of the original dataset.

## Code availability

All the code supporting this work will be available at the GitHub repository after acceptance of manuscript.

## Reference

1  Pedrosa, I., Suárez-Álvarez, J., Lozano, L. M., Muñiz, J. & García-Cueto, E. J. J. o. P. A. Assessing perceived emotional intelligence in adolescents: new validity evidence of trait Meta-Mood Scale–24. *Journal of Psychoeducational Assessment* **32**, 737-746, doi:10.1177/0734282914539238 (2014).

2  Orth, Z. & van Wyk, B. Adolescent mental wellness: a systematic review protocol of instruments measuring general mental health and well-being. *BMJ open* **10**, e037237, doi:10.1136/bmjopen-2020-037237 (2020).

3  Kim-Cohen, J. *et al.* Prior juvenile diagnoses in adults with mental disorder: developmental follow-back of a prospective-longitudinal cohort. *Archives of general psychiatry* **60**, 709-717, doi:10.1001/archpsyc.60.7.709 (2003).

4  Kessler, R. C. *et al.* Lifetime prevalence and age-of-onset distributions of mental disorders in the World Health Organization's World Mental Health Survey Initiative. *World psychiatry : official journal of the World Psychiatric Association (WPA)* **6**, 168-176 (2007).

5  Keeley, B. The State of the World's Children 2021: On My Mind--Promoting, Protecting and Caring for Children's Mental Health. *UNICEF* (2021).

6  Kieling, C. *et al.* Child and adolescent mental health worldwide: evidence for action. *Lancet (London, England)* **378**, 1515-1525, doi:10.1016/s0140-6736(11)60827-1 (2011).

7  Delamater, A. M., Guzman, A. & Aparicio, K. Mental health issues in children and adolescents with chronic illness. *International Journal of Human Rights in Healthcare* **10**, 163-173, doi:10.1108/IJHRH-05-2017-0020 (2017).

8  Costello, E. J., He, J. P., Sampson, N. A., Kessler, R. C. & Merikangas, K. R. Services for adolescents with psychiatric disorders: 12-month data from the National Comorbidity Survey-Adolescent. *Psychiatric services (Washington, D.C.)* **65**, 359-366, doi:10.1176/appi.ps.201100518 (2014).

9  Haberer, J. E., Trabin, T. & Klinkman, M. Furthering the reliable and valid measurement of mental health screening, diagnoses, treatment and outcomes through health information technology. *General hospital psychiatry* **35**, 349-353, doi:10.1016/j.genhosppsych.2013.03.009 (2013).

10 Castiglioni, M. & Laudisa, F. Toward psychiatry as a 'human' science of mind. The case of depressive disorders in DSM-5. *Frontiers in psychology* **5**, 1517, doi:10.3389/fpsyg.2014.01517 (2014).

11 Fakhoury, M. Artificial Intelligence in Psychiatry. *Advances in experimental medicine and biology* **1192**, 119-125, doi:10.1007/978-981-32-9721-0_6 (2019).

12 Aguirre Velasco, A., Cruz, I. S. S., Billings, J., Jimenez, M. & Rowe, S. What are the barriers, facilitators and interventions targeting help-seeking behaviours for common mental health problems in adolescents? A systematic review. *BMC Psychiatry* **20**, 293, doi:10.1186/s12888-020-02659-0 (2020).

13 Rasouli, S., Gupta, G., Ghafurian, M. & Dautenhahn, K. Proposed Applications of Social Robots in Interventions for Children and Adolescents with Social Anxiety. *Sixteenth*

*International Conference on Tangible, Embedded, and Embodied Interaction*, Article 71, doi:10.1145/3490149.3505575 (2022).

14  Abbasi, N. I. *et al.* Can Robots Help in the Evaluation of Mental Wellbeing in Children? An Empirical Study. *2022 31st IEEE International Conference on Robot and Human Interactive Communication (RO-MAN)*, 1459-1466, doi:10.1109/RO-MAN53752.2022.9900843 (2022).

15  Richmond-Rakerd, L. S., D'Souza, S., Milne, B. J., Caspi, A. & Moffitt, T. E. Longitudinal Associations of Mental Disorders With Physical Diseases and Mortality Among 2.3 Million New Zealand Citizens. *JAMA network open* **4**, e2033448, doi:10.1001/jamanetworkopen.2020.33448 (2021).

16  McGorry, P. D. *et al.* Designing and scaling up integrated youth mental health care. *World psychiatry : official journal of the World Psychiatric Association (WPA)* **21**, 61-76, doi:10.1002/wps.20938 (2022).

17  Wu, Z. *et al.* Changes of psychotic-like experiences and their association with anxiety/depression among young adolescents before COVID-19 and after the lockdown in China. *Schizophrenia Research* **237**, 40-46, doi:10.1016/j.schres.2021.08.020 (2021).

18  Wang, J., Li, Y. & He, E. J. P. S. Development and standardization of mental health scale for middle school students in China. *Psychosoc. Sci* **4**, 15-20 (1997).

19  Dong, R.-b. & Dou, K.-y. Changes in physical activity level of adolescents and its relationship with mental health during regular COVID-19 prevention and control. *Brain and Behavior* **n/a**, e3116, doi:https://doi.org/10.1002/brb3.3116 (2023).

20  Desideri, L. *et al.* Using a Humanoid Robot as a Complement to Interventions for Children with Autism Spectrum Disorder: a Pilot Study. *Advances in Neurodevelopmental Disorders* **2**, 273-285, doi:10.1007/s41252-018-0066-4 (2018).

21  Alves-Oliveira, P. *et al.* Robot-mediated interventions for youth mental health. *Design for Health* **6**, 138-162, doi:10.1080/24735132.2022.2101825 (2022).

22  Li, T. W. *et al.* Tell Me About It: Adolescent Self-Disclosure with an Online Robot for Mental Health. *Companion of the 2023 ACM/IEEE International Conference on Human-Robot Interaction*, 183–187, doi:10.1145/3568294.3580068 (2023).

23  Doraiswamy, P. M., Blease, C. & Bodner, K. Artificial intelligence and the future of psychiatry: Insights from a global physician survey. *Artif Intell Med* **102**, 101753, doi:10.1016/j.artmed.2019.101753 (2020).

24  Dwyer, D. & Koutsouleris, N. Annual Research Review: Translational machine learning for child and adolescent psychiatry. *Journal of Child Psychology and Psychiatry* **63**, 421-443, doi:10.1111/jcpp.13545 (2022).

25  Moura, I. *et al.* Digital Phenotyping of Mental Health using multimodal sensing of multiple situations of interest: A Systematic Literature Review. *Journal of Biomedical Informatics* **138**, 104278, doi:10.1016/j.jbi.2022.104278 (2023).

26  Fan, D. *et al.* Self-shrinking soft demoulding for complex high-aspect-ratio microchannels. *Nature Communications* **13**, 5083, doi:10.1038/s41467-022-32859-z (2022).

27  Coutts, L. V., Plans, D., Brown, A. W. & Collomosse, J. J. J. o. B. I. Deep learning with wearable based heart rate variability for prediction of mental and general health. *J Biomed Inform* **112**, 103610, doi:10.1016/j.jbi.2020.103610 (2020).

28  Arya, L. & Sethia, D. HRV and GSR as Viable Physiological Markers for Mental Health

Recognition. *2022 14th International Conference on COMmunication Systems & NETworkS (COMSNETS)*, 37-42 (2022).

29  Tiwari, S. & Agarwal, S. J. B. D. A Shrewd Artificial Neural Network-Based Hybrid Model for Pervasive Stress Detection of Students Using Galvanic Skin Response and Electrocardiogram Signals. *Big Data* **9**, 427-442, doi:10.1089/big.2020.0256 (2021).

30  Tariq, Q. *et al.* Mobile detection of autism through machine learning on home video: A development and prospective validation study. *PLoS medicine* **15**, e1002705, doi:10.1371/journal.pmed.1002705 (2018).

31  Cohen, J. *et al.* A feasibility study using a machine learning suicide risk prediction model based on open-ended interview language in adolescent therapy sessions. *Int J Environ Res Public Health* **17**, 8187, doi:10.3390/ijerph17218187 (2020).

32  Saha, K., Yousuf, A., Boyd, R. L., Pennebaker, J. W. & De Choudhury, M. J. S. r. Social media discussions predict mental health consultations on college campuses. *Sci Rep* **12**, 123, doi:10.1038/s41598-021-03423-4 (2022).

33  Huang, Y. *et al.* What Makes Multi-Modal Learning Better than Single (Provably). *Advances in Neural Information Processing Systems*, 10944-10956, doi:10.48550/arXiv.2106.04538 (2021).

34  Tunc, B. *et al.* Diagnostic shifts in autism spectrum disorder can be linked to the fuzzy nature of the diagnostic boundary: a data-driven approach. *Journal of Child Psychology and Psychiatry* **62**, 1236-1245, doi:10.1111/jcpp.13406 (2021).

35  Zhang-James, Y. *et al.* Machine-Learning prediction of comorbid substance use disorders in ADHD youth using Swedish registry data. *Journal of Child Psychology and Psychiatry* **61**, 1370-1379, doi:10.1111/jcpp.13226 (2020).

36  Worthington, M. A. *et al.* Individualized Prediction of Prodromal Symptom Remission for Youth at Clinical High Risk for Psychosis. *Schizophrenia Bulletin* **48**, 395-404, doi:10.1093/schbul/sbab115 (2022).

37  Marti-Puig, P., Capra, C., Vega, D., Llunas, L. & Sole-Casals, J. A Machine Learning Approach for Predicting Non-Suicidal Self-Injury in Young Adults. *Sensors* **22**, doi:10.3390/s22134790 (2022).

38  Choi, J.-H. & Lee, J.-S. EmbraceNet: A robust deep learning architecture for multimodal classification. *Information Fusion* **51**, 259-270, doi:doi.org/10.1016/j.inffus.2019.02.010 (2019).

39  Hughes, C. W. & Melson, A. G. in *Handbook of psychiatric measures, 2nd ed.*    251-308 (American Psychiatric Publishing, Inc., 2008).

40  Gangemi, A., Tenore, K. & Mancini, F. Two Reasoning Strategies in Patients With Psychological Illnesses. *Frontiers in psychology* **10**, 2335, doi:10.3389/fpsyg.2019.02335 (2019).

41  Johnson-Laird, P. N., Mancini, F. & Gangemi, A. A hyper-emotion theory of psychological illnesses. *Psychological review* **113**, 822-841, doi:10.1037/0033-295x.113.4.822 (2006).

42  Schumann, G. *et al.* The IMAGEN study: reinforcement-related behaviour in normal brain function and psychopathology. *Molecular Psychiatry* **15**, 1128-1139, doi:10.1038/mp.2010.4 (2010).

43  Karcher, N. R. & Barch, D. M. The ABCD study: understanding the development of risk for mental and physical health outcomes. *Neuropsychopharmacology* **46**, 131-142,

doi:10.1038/s41386-020-0736-6 (2021).

44  Zhang, L. *et al.* AI-boosted CRISPR-Cas13a and total internal reflection fluorescence microscopy system for SARS-CoV-2 detection.  **3**, doi:10.3389/fsens.2022.1015223 (2022).

45  Liu, Y. *et al.* Mixed-UNet: Refined class activation mapping for weakly-supervised semantic segmentation with multi-scale inference.  **4**, doi:10.3389/fcomp.2022.1036934 (2022).

46  Rezapour, M. & Hansen, L. A machine learning analysis of COVID-19 mental health data. *Scientific Reports* **12**, 14965, doi:10.1038/s41598-022-19314-1 (2022).

47  Mohamed, E. S. *et al.* A hybrid mental health prediction model using Support Vector Machine, Multilayer Perceptron, and Random Forest algorithms. *Healthcare Analytics* **3**, 100185 (2023).

48  Garriga, R. *et al.* Machine learning model to predict mental health crises from electronic health records. *Nature Medicine* **28**, 1240-1248, doi:10.1038/s41591-022-01811-5 (2022).

49  Montgomery, J., Hendry, J., Wilson, J. A., Deary, I. J. & MacKenzie, K. Pragmatic detection of anxiety and depression in a prospective cohort of voice outpatient clinic attenders. *Clinical Otolaryngology* **41**, 2-7, doi:10.1111/coa.12459 (2016).

50  Mo, L., Li, H. & Zhu, T. Exploring the Suicide Mechanism Path of High-Suicide-Risk Adolescents-Based on Weibo Text Analysis. *Int J Environ Res Public Health* **19**, doi:10.3390/ijerph191811495 (2022).

51  Zhang, R. *et al.* RCMNet: A deep learning model assists CAR-T therapy for leukemia. *Computers in biology and medicine* **150**, 106084, doi:https://doi.org/10.1016/j.compbiomed.2022.106084 (2022).

52  Bhardwaj, V. *et al.* Machine Learning for Endometrial Cancer Prediction and Prognostication. *Frontiers in oncology* **12**, 852746, doi:10.3389/fonc.2022.852746 (2022).

53  Xie, Y. *et al.* Stroke prediction from electrocardiograms by deep neural network. *Multimedia Tools and Applications* **80**, 17291-17297, doi:10.1007/s11042-020-10043-z (2021).

54  Githinji, B. *et al.* Multidimensional Hypergraph on Delineated Retinal Features for Pathological Myopia Task. *25th International Conference on Medical Image Computing and Computer Assisted Intervention (MICCAI)* **13432**, 550-559, doi:10.1007/978-3-031-16434-7_53 (2022).

55  Li, F. *et al.* Therapeutic effect of ketogenic diet treatment on type 2 diabetes. *Journal of Future Foods* **2**, 177-183, doi:https://doi.org/10.1016/j.jfutfo.2022.03.004 (2022).

56  Yang, J. *et al.* Prediction of HER2-positive breast cancer recurrence and metastasis risk from histopathological images and clinical information via multimodal deep learning. *Computational and Structural Biotechnology Journal* **20**, 333-342, doi:10.1016/j.csbj.2021.12.028 (2022).

57  Arbabshirani, M. R., Plis, S., Sui, J. & Calhoun, V. D. Single subject prediction of brain disorders in neuroimaging: Promises and pitfalls. *Neuroimage* **145**, 137-165, doi:10.1016/j.neuroimage.2016.02.079 (2017).

58  Lei, Z. *et al.* Detection of Frog Virus 3 by Integrating RPA-CRISPR/Cas12a-SPM with Deep Learning. *ACS Omega* **8**, 32555-32564, doi:10.1021/acsomega.3c02929 (2023).

59   Zhang, Y. *et al.* Weighted Convolutional Motion-Compensated Frame Rate Up-Conversion Using Deep Residual Network. *IEEE Transactions on Circuits and Systems for Video Technology* **30**, 11-22, doi:10.1109/TCSVT.2018.2885564 (2020).

60   Ashwini, B. Robot Assisted Diagnosis of Autism in Children. *Proceedings of the 2020 International Conference on Multimodal Interaction*, 728–732, doi:10.1145/3382507.3421162 (2020).

61   Uban, A.-S., Chulvi, B. & Rosso, P. Explainability of Depression Detection on Social Media: From Deep Learning Models to Psychological Interpretations and Multimodality. *Early Detection of Mental Health Disorders by Social Media Monitoring: The First Five Years of the eRisk Project*, 289-320, doi:10.1007/978-3-031-04431-1_13 (2022).

62   Stone, L. L., Otten, R., Engels, R. C. M. E., Vermulst, A. A. & Janssens, J. M. A. M. Psychometric Properties of the Parent and Teacher Versions of the Strengths and Difficulties Questionnaire for 4- to 12-Year-Olds: A Review. *Clinical child and family psychology review* **13**, 254-274, doi:10.1007/s10567-010-0071-2 (2010).

63   Castellanos-Ryan, N., O'Leary-Barrett, M., Sully, L. & Conrod, P. Sensitivity and Specificity of a Brief Personality Screening Instrument in Predicting Future Substance Use, Emotional, and Behavioral Problems: 18-Month Predictive Validity of the Substance Use Risk Profile Scale. *Alcoholism: Clinical and Experimental Research* **37**, E281-E290, doi:https://doi.org/10.1111/j.1530-0277.2012.01931.x (2013).

64   Gili, M. *et al.* Mental disorders as risk factors for suicidal behavior in young people: A meta-analysis and systematic review of longitudinal studies. *Journal of Affective Disorders* **245**, 152-162, doi:doi.org/10.1016/j.jad.2018.10.115 (2019).

65   Hirschfeld, R. M. The Comorbidity of Major Depression and Anxiety Disorders: Recognition and Management in Primary Care. *Primary care companion to the Journal of clinical psychiatry* **3**, 244-254, doi:10.4088/pcc.v03n0609 (2001).

66   Davids, A. & Parenti, A. N. Time orientation and interpersonal relations of emotionally disturbed and normal children. *The Journal of Abnormal and Social Psychology* **57**, 299-305, doi:10.1037/h0047687 (1958).

67   Chaddad, A., Peng, J., Xu, J. & Bouridane, A. Survey of Explainable AI Techniques in Healthcare. *Sensors (Basel, Switzerland)* **23**, doi:10.3390/s23020634 (2023).

68   Duijndam, S., Karreman, A., Denollet, J. & Kupper, N. Physiological and emotional responses to evaluative stress in socially inhibited young adults. *Biological Psychology* **149**, doi:10.1016/j.biopsycho.2019.107811 (2020).

69   Byck, G. R. *et al.* Effect of housing relocation and neighborhood environment on adolescent mental and behavioral health. *Journal of Child Psychology and Psychiatry* **56**, 1185-1193, doi:10.1111/jcpp.12386 (2015).

70   Gross, J. J. & Levenson, R. W. EMOTION ELICITATION USING FILMS. *Cognition & Emotion* **9**, 87-108, doi:10.1080/02699939508408966 (1995).

71   Schaefer, A., Nils, F., Sanchez, X. & Philippot, P. Assessing the effectiveness of a large database of emotion-eliciting films: A new tool for emotion researchers. *Cognition & Emotion* **24**, 1153-1172, doi:10.1080/02699930903274322 (2010).

72   Ma, M. *et al.* Smil: Multimodal learning with severely missing modality. *Proceedings of the AAAI Conference on Artificial Intelligence* **35**, 2302-2310 (2021).

73   Dolci, G. *et al.* in *2023 IEEE International Conference on Acoustics, Speech, and Signal*

| | |
|---|---|
| | *Processing Workshops (ICASSPW)*.   1-5. |
| 74 | Hillege, R. H. L., Lo, J. C., Janssen, C. P. & Romeijn, N. The Mental Machine: Classifying Mental Workload State from Unobtrusive Heart Rate-Measures Using Machine Learning. *Adaptive Instructional Systems. Second International Conference, AIS 2020. Held as Part of the 22nd HCI International Conference, HCII 2020. Proceedings. Lecture Notes in Computer Science (LNCS 12214)*, 330-349, doi:10.1007/978-3-030-50788-6_24 (2020). |
| 75 | Davila, M. I., Lewis, G. F. & Porges, S. W. The PhysioCam: A Novel Non-Contact Sensor to Measure Heart Rate Variability in Clinical and Field Applications. *Frontiers in Public Health* **5**, doi:10.3389/fpubh.2017.00300 (2017). |
| 76 | Nishikawa, M. *et al.* in *43rd Annual International Conference of the IEEE-Engineering-in-Medicine-and-Biology-Society (IEEE EMBC)*.   7016-7019 (2021). |
| 77 | Camras, L. EMOTION - A PSYCHOEVOLUTIONARY SYNTHESIS - PLUTCHIK,R. *American Journal of Psychology* **93**, 751-753, doi:10.2307/1422394 (1980). |
| 78 | Zhu, H. B., Han, G. J., Shu, L. & Zhao, H. ArvaNet: Deep Recurrent Architecture for PPG-Based Negative Mental-State Monitoring. *Ieee Transactions on Computational Social Systems* **8**, 179-190, doi:10.1109/tcss.2020.2977715 (2021). |
| 79 | Bonicatto, S., Dew, M. A., Soria, J. J. & Seghezzo, M. E. Validity and reliability of Symptom Checklist '90 (SCL90) in an Argentine population sample. *Social Psychiatry and Psychiatric Epidemiology* **32**, 332-338, doi:10.1007/bf00805438 (1997). |
| 80 | Luo, Y. *et al.* Mental Health Problems and Associated Factors in Chinese High School Students in Henan Province: A Cross-Sectional Study. *Int J Environ Res Public Health* **17**, doi:10.3390/ijerph17165944 (2020). |
| 81 | Chen, Y. *et al.* Social Identity, Core Self-Evaluation, School Adaptation, and Mental Health Problems in Migrant Children in China: A Chain Mediation Model. *Int J Environ Res Public Health* **19**, doi:10.3390/ijerph192416645 (2022). |
| 82 | das, A., Jena, M. R. & Barik, K. K. Mel-Frequency Cepstral Coefficient (MFCC) - a Novel Method for Speaker Recognition. *Digital Technologies* **1**, 1-3, doi:10.12691/dt-1-1-1 (2014). |
| 83 | Baevski, A., Zhou, H., Mohamed, A. & Auli, M. wav2vec 2.0: a framework for self-supervised learning of speech representations. *Proceedings of the 34th International Conference on Neural Information Processing Systems*, Article 1044 (2020). |
| 84 | Hsu, W.-N. *et al.* Robust wav2vec 2.0: Analyzing domain shift in self-supervised pre-training. *arXiv preprint*, doi:arXiv:2104.01027 (2021). |
| 85 | Vaswani, A. *et al.* Attention is all you need. *Proceedings of the 31st International Conference on Neural Information Processing Systems*, 6000–6010 (2017). |
| 86 | Yinhan, L. *et al.* RoBERTa: A Robustly Optimized BERT Pretraining Approach. *arXiv preprint*, 13 pp.-13 pp., doi:arXiv:1907.11692 (2019). |
| 87 | Cui, Y., Yang, Z. & Liu, T. PERT: Pre-training BERT with Permuted Language Model. *arXiv preprint*, doi:arXiv:2203.06906 (2022). |
| 88 | Devlin, J., Chang, M.-W., Lee, K. & Toutanova, K. BERT: Pre-training of Deep Bidirectional Transformers for Language Understanding. *arXiv preprint*, 4171-4186, doi:arXiv:1810.04805 (2018). |
| 89 | Kartynnik, Y., Ablavatski, A., Grishchenko, I. & Grundmann, M. Real-time facial surface geometry from monocular video on mobile GPUs. *arXiv preprint*, |

90  Ablavatski, A., Vakunov, A., Grishchenko, I., Raveendran, K. & Zhdanovich, M. J. a. p. a. Real-time Pupil Tracking from Monocular Video for Digital Puppetry. *arXiv preprint*, doi:arXiv:2006.11341 (2020).

89 ...doi:arXiv:1907.06724 (2019).

91  Zheng, W.-L., Liu, W., Lu, Y., Lu, B.-L. & Cichocki, A. EmotionMeter: A Multimodal Framework for Recognizing Human Emotions. *Ieee Transactions on Cybernetics* **49**, 1110-1122, doi:10.1109/tcyb.2018.2797176 (2019).

92  Zekveld, A. A., Heslenfeld, D. J., Johnsrude, I. S., Versfeld, N. J. & Kramer, S. E. The eye as a window to the listening brain: Neural correlates of pupil size as a measure of cognitive listening load. *Neuroimage* **101**, 76-86, doi:10.1016/j.neuroimage.2014.06.069 (2014).

93  Christ, M., Braun, N., Neuffer, J. & Kempa-Liehr, A. W. Time Series FeatuRe Extraction on basis of Scalable Hypothesis tests (tsfresh – A Python package). **307**, 72–77, doi:10.1016/j.neucom.2018.03.067 (2018).

94  Christ, M., Kempa-Liehr, A. W. & Feindt, M. J. a. p. a. Distributed and parallel time series feature extraction for industrial big data applications. *Asian Machine Learning Conference (ACML) 2016, Workshop on Learning on Big Data (WLBD), Hamilton (New Zealand)* (2016).

95  Lines, J. & Bagnall, A. Time series classification with ensembles of elastic distance measures. *Data Mining and Knowledge Discovery* **29**, 565-592, doi:10.1007/s10618-014-0361-2 (2015).